\documentclass{article}

    \PassOptionsToPackage{numbers, compress}{natbib}

\usepackage{ulem}

\usepackage[final]{neurips_2022}

\usepackage[utf8]{inputenc} 
\usepackage[T1]{fontenc}    
\usepackage{hyperref}       
\usepackage{url}            
\usepackage{booktabs}       
\usepackage{amsfonts}       
\usepackage{nicefrac}       
\usepackage{microtype}      
\usepackage{graphicx}
\usepackage{xcolor}         
\usepackage{svg}
\usepackage{amsmath,amssymb,amsthm}
\usepackage{bm,bbm}
\usepackage{dsfont}

\newcommand{\csdit}{TabCSDI}
\newcommand{\x}{\mathbf{x}}

\newcommand{\X}{\mathcal{X}}
\newcommand{\R}{\mathbb{R}}
\newcommand{\miss}{\varnothing}
\newcommand{\vecone}{\mathds{1}}
\usepackage{natbib}
\usepackage{lipsum}

\title{Diffusion models for missing value imputation in tabular data}

\author{%
  Shuhan Zheng\thanks{This work was done during his internship at Preferred Networks.} \\
  The University of Tokyo\\
   \texttt{shuhanzheng@g.ecc.u-tokyo.ac.jp} \\
   \And
  Nontawat Charoenphakdee \\
  Preferred Networks\\
  \texttt{nontawat@preferred.jp} \\
}

\begin{document}

\maketitle

\let\thefootnote\relax\footnotetext{Table Representation Learning Workshop at NeurIPS 2022}

\begin{abstract}
Missing value imputation in machine learning is the task of estimating the missing values in the dataset accurately using available information. 
In this task, several deep generative modeling methods have been proposed and demonstrated their usefulness, e.g., generative adversarial imputation networks. 
Recently, diffusion models have gained popularity because of their effectiveness in the generative modeling task in images, texts, audio, etc. 
To our knowledge, less attention has been paid to the investigation of the effectiveness of diffusion models for missing value imputation in tabular data. 
Based on recent development of diffusion models for time-series data imputation, we propose a diffusion model approach called ``Conditional Score-based Diffusion Models for Tabular data'' (\csdit).
To effectively handle categorical variables and numerical variables simultaneously, we investigate three techniques: one-hot encoding, analog bits encoding, and feature tokenization. 
Experimental results on benchmark datasets demonstrated the effectiveness of {\csdit} compared with well-known existing methods, and also emphasized the importance of the categorical embedding techniques.
\end{abstract}

\section{Introduction}

In real-world applications, it is often the case that the dataset for training a prediction model contains missing values. 
This phenomenon can happen for many reasons, e.g., human error, privacy concerns, and the difficulty of data collection. 
For example, in census data, some people might not be comfortable to reveal their sensitive information such as employment information~\citep{lillard1986we, eckert2020imputing}. 
In healthcare data, different patients may have taken different health examinations, which cause the dataset to have different available health features for each patient~\citep{wells2013strategies, hegde2019mice}.
Technically, the missing data problem can be divided into three categories: Missing completely at random (MCAR), Missing at random (MAR), Missing not at random (MNAR) (see~\citet{rubin1976inference} and~\citet{van2018flexible} for more details). 

According to~\citet{jarrett2022hyperimpute}, a missing value imputation approach can be divided into two categories.
The first category is the iterative approach. 
It is based on the idea of estimating the conditional distribution of one feature using all other available features. 
In one iteration, we will train a conditional distribution estimator to predict the value of each feature.
We will repeat the process for many iterations until the process is converged, i.e., the latest iteration does not change the prediction output significantly according to the pre-specified convergence criterion. 
This approach has been studied extensively~\citep{van2000multivariate,khan2020sice, stekhoven2012missforest, jarrett2022hyperimpute} and one of the most well-known methods is Multiple Imputation based on Chained Equations (MICE)~\citep{van2000multivariate}. 
The second category is a deep generative model approach. 
In this approach, we will train a generative model to generate values in missing parts based on observed values.
Previous methods that can be categorized in this approach are Multiple Imputation using Denoising Autoencoders (MIDA)~\citep{gondara2018mida}, Handling Incomplete Heterogeneous Data using Variational Autoencoders (HIVAE)~\citep{nazabal2020handling}, Missing Data Importance-weighted
Autoencoder (MIWAE)~\citep{mattei2019miwae}, and Generative Adversarial Imputation Nets (GAIN)~\citep{yoon2018gain}. 
Recently, diffusion model is a generative model that has demonstrated its effectiveness over other generative models in various domains, e.g., computer vision~\citep{song2019generative, ho2020denoising,croitoru2022diffusion}, time-series data~\citep{tashiro2021csdi, rasul2021autoregressive}, chemistry~\citep{luo2021predicting, xu2022geodiff}, and natural language processing~\citep{li2022diffusion, yu2022latent}. 
However, to the best of our knowledge, a diffusion model has not been proposed yet for missing value imputation in tabular data.

The goal of this paper is to develop a diffusion model approach for missing value imputation based on the recent development of diffusion model approach for missing value imputation in time-series data called CSDI~\citep{tashiro2021csdi}. 
CSDI is originally designed for time-series data and it cannot support categorical variables, which are necessary for tabular data.
To solve this problem, we propose a variant of CSDI called {\csdit} for tabular data by making it supports both the categorical and numerical features. 
Our experimental results show that {\csdit} can be successfully trained to achieve competitive performance to existing methods both in the iterative approach and generative approach. 
We can also observe that the choice of categorical embedding methods can affect performance.

\section{Problem formulation}
Let $\X =(\R \cup \{ \miss \})^d$ be an input space, where $\R$ denotes a real number space and ``$\miss$'' denotes a missing value.
In missing value imputation, we are given $d$-dimensional training dataset $\mathbf{X}_{\mathrm{tr}} = \{ \x_i \}_{i=1}^{n}$, where $n$ is the number of data. 
Without loss of generality, a feature $j \in \{1, \ldots, d\} $ of $\x_i$ is defined as $\x^j_i \in \X$, where a feature can be missing, numerical variable, or categorical variable.  
This paper focuses on an inductive setting where the goal is to find an imputation function $f \colon \X \to \R^d$ that transforms the input that allows missing value $\X$ to the $d$-dimension real values $\R^d$.
A desirable $f$ should be able to replace the missing values with reasonable values. 

To evaluate the performance of $f$, we are given test input data $\mathbf{X}_{\mathrm{te}} = \{ \x_i \}_{i=1}^{n}$ and ground truths $\mathbf{Y}_{\mathrm{te}} = \{ y^j_i \in \R \colon \x^j_i = \miss$\}.
For $\x^j_i$, we define $\widehat{\x}^j_i$ to be an imputed feature obtained from $f(\x_i)$ for a feature $j$. 
Let $M^j = \{ i: \x^j_i=\miss \}$ be the set of missing value indices and $N^j_\mathrm{miss} = |M^j|$ be the number of missing values for a feature $j$. 
To calculate the error of $f$, we use the root mean squared error (RMSE) if $j$ is numerical and the error rate (Err) if $j$ is categorical:
\begin{align*}
    \mathrm{RMSE}(j) = \sqrt{\frac{\sum_{i \in M^j} (\widehat{\x}^j_i - y^j_i)^2 } {N^j_\mathrm{miss}}}, \quad \mathrm{Err}(j) = \frac{1}{N^j_\mathrm{miss}} \sum_{i \in M^j}  \vecone_{[\widehat{\x}^j_i \neq y^j_i]},
\end{align*}
where $\vecone_{[\cdot]}$ is an indicator function that returns $1$ if the condition holds and $0$ otherwise.


\section{\csdit: Conditional Score-based Diffusion Models for Tabular data}
In this section, we describe our proposed diffusion model method for missing value imputation in tabular data by describing CSDI~\citep{tashiro2021csdi} and how to modify it for handling tabular data.
\subsection{Conditional Score-based Diffusion Model (CSDI)}
Diffusion model contains two processes: the forward noising process where we iteratively inject the noise into the input data, and the reverse denoising process where we iteratively denoise the data.
In the standard training process of diffusion model~\citep{song2019generative, ho2020denoising}, only the reverse process requires training while the forward process is always fixed. 
We omit the details of diffusion model for brevity (see \citet{song2019generative, ho2020denoising} for more information).

Based on the idea of diffusion model,~\citet{tashiro2021csdi} recently proposed a diffusion model called CSDI for missing value imputation for time-series data. 
The key idea of CSDI can be explained as follows.
Instead of reconstructing the whole input $\x$ by straightforwardly using the diffusion model, aka., unconditional diffusion model (see Appendix C of~\citet{tashiro2021csdi}), CSDI separates input $\x$ into two parts: the observed part (aka., conditional part) $\x^{co}$ and the unobserved part to predict (aka., target part) $\x^{ta}$. The goal of the diffusion model is to model the following distribution:
\begin{equation*}
p_{\theta}(\mathbf{x}^{ta}_{t-1}|\mathbf{x}^{ta}_{t}, \mathbf{x}^{co}_{0}) = \mathcal{N}(\mathbf{x}_{t-1}^{ta};\mathbf{\mu}_\theta(\mathbf{x}_{t}^{ta},t|\mathbf{x}^{co}_{0}), \sigma \mathbf{I}),
\end{equation*} 
where $t \in \{1, \ldots, T\}$ denotes the iteration round of the process and $T$ is a hyperparameter. 
We need to model $\mathbf{\mu}_\theta$ that focuses only on predicting the values of the unobserved part. It is observed that the conditional diffusion model can achieve better performance than the unconditional one.

In our study, we followed the formulation of~\citet{tashiro2021csdi} as our objective function.
For the architecture part, we slightly modified the architecture proposed in CSDI to appropriately handle tabular data. 
More specifically, we removed the temporal transformer layer of the original CSDI architecture since our data does not contain temporal information and use a simple residual connection of transformer encoder and multi-layered perceptron.




\subsection{Handling categorical variables}
\begin{figure}
  \centering
  \includegraphics[scale=0.5]{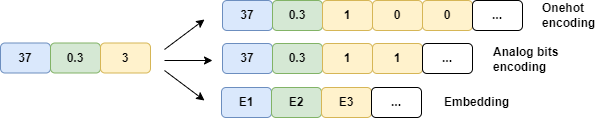}
  \caption{Example of handling categorical variables in one-hot encoding, analog bits encoding, and embeddings. The categorical variable is marked as the yellow block. Two numerical variables are marked as blue and green blocks.}
  \label{fig:encoding}
  \vspace{-0.2in}
\end{figure}

In the original CSDI, it is assumed that the input features contain only numerical variables, which is not the case for tabular data. 
In this section, we extend CSDI to support categorical variables by proposing three different techniques: (1) one-hot encoding, (2) analog bits encoding, and (3) feature tokenization. 
Figure~\ref{fig:encoding} illustrates how each encoding works. 
The categorical variable is marked as yellow and we assume that there are three different categories for this feature. 
Without loss of generality, for one-hot encoding, the representation can be $[1,0,0]$. 
For analog bits, we follow the encoding scheme proposed by~\citet{chen2022analog}. 
In our example, the categorical variable will take two columns and represented in binary bits as $[1,1]$.  
To make the data more distinguishable, we further convert $0$ to $-1$ in one-hot and analog bits encoding. 
For feature tokenizer~\citep{gorishniy2021revisiting}, we transform both numerical and categorical variables together to embeddings. 
In our example, variables will have embedding vectors with the same length, i.e., $E_1, E_2, E_3 \in \R^e$. In sum, analog bits encoding takes less columns compared to one-hot encoding but will make the encoded vector complex. 
Feature tokenizer lets all variables have embeddings with the same length.

Then, we train the model with the processed input. 
After obtaining the raw output, different handling schemes require different recover procedures. 
For one-hot encoding, we treat the index of the largest element as the model inferred category. 
For analog bits encoding, we convert every output element to $1$ if the output element is larger than $0$, otherwise we convert it to $-1$.
In the FT scheme, we need to recover both numerical and categorical variables back from the embeddings~\citep{gorishniy2021revisiting}. 
For numerical variables, we divide the diffusion model output by the corresponding embedding element-wise and use the average value as the final model output. 
For categorical variables, we calculate the Euclidean distance between {\csdit} outputs and every categorical embedding and set the category of closest embedding (i.e., $1$-nearest neighbor) as the final model output.

\section{Experimental results}
In this section, we report experiments on pure numerical datasets and mixed variable datasets to show the effectiveness of \csdit.

\textbf{Datasets:}
We used seven datasets.
Census Income Data Set (Census), Wine Quality (Wine), Concrete Compressive Strength (Concrete), Libras Movement (Libras) and Breast Cancer Wisconsin (Breast) were obtained from UCI Machine Learning Repository~\citep{Dua:2019}. 
COVID-19\footnote{https://www.kaggle.com/datasets/tanmoyx/covid19-patient-precondition-dataset} and Diabetes\footnote{https://www.kaggle.com/datasets/alexteboul/diabetes-health-indicators-dataset} were obtained from Kaggle. 
Dataset information is detailed in Appendix~\ref{sec:app-dataset}.
Note that Diabetes and COVID-19 datasets only have binary category variables and we preprocessed the numerical variables for all datasets by min-max normalization. 

\textbf{Comparison methods:}
In our experiments, we compare our proposed method with a simple baseline that uses training data's mean values for numerical variables and mode values for categorical variables (Mean / Mode).
We used MICE method with linear regression and logistic regression (MICE (linear)) and MICE method based on random forest (MissForest).
We also used GAIN as a representative method for a deep generative model approach.
The code implementation for MICE (linear), MissForest, and GAIN was provided by the Hyperimpute framework~\citep{jarrett2022hyperimpute}\footnote{https://github.com/vanderschaarlab/hyperimpute}. 
For {\csdit}, we built our code based on CSDI~\citep{tashiro2021csdi}\footnote{https://github.com/ermongroup/CSDI}.
Hyperparameter information is detailed in Appendix~\ref{sec:app-hyperparams}.

\textbf{Results:} First, we show our results on three mixed variable datasets (Diabetes, Census and COVID-19). 
The following Table~\ref{table:mixed_resutls} shows the comparison between different imputation methods and categorical variable handling schemes. 
Our proposed methods ({\csdit}) reached the lowest RMSE in Diabetes and Census datasets. 
MissForest reached the lowest error rate in the Diabetes and Census datasets. 
The RMSE difference between three categorical handling methods was not evident.
However, {\csdit} with FT obtained the lowest error rate in the Census dataset compared to other two categorical handling methods, where the analog bits approach is superior to one-hot. 
Second, we show our results on four pure numerical datasets in Table~\ref{table:numerical_resutls}. 
It can be observed that our proposed {\csdit} has best performance against other comparison methods for three out of four datasets.

\begin{table}
\centering
\caption{RMSE and error rate performance for comparison methods on three mixed variable datasets. 
Note that one-hot and analog bits are equivalent for a dataset without multi-categorical variables.}
\label{table:mixed_resutls}
\scalebox{0.8}{
\begin{tabular}{lllllll}
\hline
                      & \multicolumn{2}{c}{Diabetes}                    & \multicolumn{2}{c}{COVID-19}                       & \multicolumn{2}{c}{Census}                      \\ \cline{2-7} 
                      & RMSE                   & Error rate             & RMSE                   & Error rate             & RMSE                   & Error rate             \\ \hline
Mean / Mode           & 0.222 (0.003)          & 0.260 (0.004)          & 0.138 (0.002)          & 0.144 (0.002)          & 0.120 (0.003)          & 0.424 (0.003)          \\
MICE (linear)                  & 0.263 (0.002)          & 0.270 (0.004)          & 0.125 (0.003)          & 0.300 (0.038)          & 0.101 (0.002)          & 0.530 (0.011)          \\
MissForest            & 0.216 (0.003)          & \textbf{0.214 (0.001)} & \textbf{0.120 (0.002)} & 0.131 (0.002)          & 0.112 (0.004)          & \textbf{0.300 (0.014)} \\
GAIN                  & 0.202 (0.003)          & 0.282 (0.005)          & 0.127 (0.002)          & 0.217 (0.011)          & 0.123 (0.057)          & 0.412 (0.012)          \\
\csdit / one-hot     & \textbf{0.197 (0.001)} & 0.222 (0.005)          & 0.122 (0.003)          & 0.111 (0.012)          & 0.099 (0.004)          & 0.400 (0.033)          \\
\csdit / analog bits & \textbf{0.197 (0.001)} & 0.222 (0.005)          & 0.122 (0.003)          & 0.111 (0.012)          & 0.103 (0.004)          & 0.376 (0.013)          \\
\csdit / FT          & 0.206 (0.002)          & 0.224 (0.004)          & 0.123 (0.002)          & \textbf{0.107 (0.002)} & \textbf{0.098 (0.003)} & 0.345 (0.002)          \\ \hline
\end{tabular}}
\end{table}

\begin{table}
\caption{RMSE performance of comparison methods on four pure numerical datasets.}
\label{table:numerical_resutls}
\centering
\begin{tabular}{lllll}
\hline
Methods    & Wine           & Concrete       & Libras        & Breast         \\ \hline
Mean       & 0.076 (0.003)  & 0.217 (0.007)  & 0.099 (0.001) & 0.263 (0.009)  \\
MICE (linear)     & 0.065 (0.003)  & 0.153 (0.006)  & 0.034 (0.001) & 0.154  (0.011) \\
MissForest & \textbf{0.060 (0.002)} & 0.173 (0.005)  & 0.024 (0.001) & 0.163 (0.014)  \\
GAIN       & 0.072 (0.004)  & 0.203  (0.007) & 0.089 (0.006) & 0.165 (0.006)  \\
\csdit    & 0.065 (0.004) & \textbf{0.131 (0.008)}  & \textbf{0.011 (0.001)} & \textbf{0.153 (0.003)}  \\ \hline
\end{tabular}
\vspace{-0.2in}
\end{table}
\textbf{Discussions:} Based on the results, {\csdit} is observed to be effective in imputing numerical variables, where it obtained the best RMSE performance for 5 out of 7 datasets. 
Different from previous generative models, the diffusion model performs decoding through a reverse process. 
{\csdit} can benefit from this iterative approximation reverse process, which allows the neural network to gradually figure out the target value. 
Moreover, our results suggest the effectiveness of FT in handling categorical variables. The superiority of FT is evident in the Census dataset (the only multi-category mixed data types dataset). 
One possible reason is that FT treats all variables equally. 
That is, all numerical variables will have embedding vectors with the same length.
This strategy avoids the problem of column imbalance. 
Column imbalance can happen in one-hot and analog bits encoding, where the more categories the category variable contain, the more columns it will take.

\section{Conclusions and future work}
We have proposed a diffusion model-based method for missing value imputation called {\csdit}. 
We demonstrated that {\csdit} can obtain competitive performance with other well-known imputation methods. 
Particularly, {\csdit} works well for numerical variables imputation. 
We also explored different schemes for handling categorical variables and found that FT embedding gives evident better performance compared to one-hot encoding and analog bits in the Census dataset. 
Future work for {\csdit} that can be considered are (1) investigation of the inference time,  (2) model architecture improvement, and (3) theoretical analysis of the loss function.

\section*{Acknowledgements}
The authors would like to thank Shin-ichi Maeda, Kohei Hayashi, and Kenta Oono for helpful discussions during the PFN summer internship program 2022.


\bibliography{references}
\newpage
\appendix

\section{Dataset information}\label{sec:app-dataset}
Table~\ref{table:datasets} shows the characteristic of each dataset. 

\begin{table}[htbp]
\caption{Dataset information}
\centering
\begin{tabular}{llll}
\hline
Datasets & \# Categorical & \# Numerical & \# Samples     \\ \hline
Census   & 9              & 6            & 20000 \\
Diabetes & 15             & 7            & 20000  \\
COVID-19    & 18             & 1            & 56660          \\
Wine     & 0              & 12           & 4898           \\
Concrete & 0              & 9            & 1030           \\
Libras   & 0              & 91           & 360            \\
Breast   & 0              & 10           & 699            \\ \hline
\end{tabular}
\label{table:datasets}
\end{table}

\section{Hyperparameters}\label{sec:app-hyperparams}
We did not heavily tune parameters for our model as well as the baseline models. 
We used default settings for most hyperparameters for all experiments except the batch size and the number of epochs, which depends on the dataset size. 
For {\csdit}, We set the number of layers to 2 for the COVID-19 dataset and 4 for other datasets. 
We set the initial learning rate as 0.0005. We use Adam optimizer with MultiStepLR with 0.1 decay at 25\%, 50\%, 75\%, and 90\% of the total epochs. 
The number of channels is set as 64. 
The number of heads in the transformer encoder is set as 4. The dimension of the diffusion embedding and feature embeddings is 128 and 64, separately. 
The number of reverse steps is 150 for Diabetes, COVID-19, and all numerical datasets, and 100 for the Census dataset.

\end{document}